\pdfoutput=1

\documentclass[11pt]{article}

\usepackage[]{acl}
\usepackage{times}
\usepackage{latexsym}
\usepackage{amsmath}
\usepackage{enumitem}
\usepackage{CJKutf8}
\usepackage{booktabs}
\usepackage{arydshln}
\usepackage{nicematrix,tikz}

\NiceMatrixOptions
  {
    custom-line = 
     {
       letter = : ,
       command = dashedline , 
       ccommand = cdashedline ,
       tikz = dashed
     }
  }

\usepackage[T1]{fontenc}

\usepackage[utf8]{inputenc}

\usepackage{microtype}

\usepackage{inconsolata}
\usepackage{comment}
\makeatother

\makeatletter
\newcommand{\thickhline}{%
    \noalign {\ifnum 0=`}\fi \hrule height 1pt
    \futurelet \reserved@a \@xhline
}
\makeatother

\newcommand{\ours}{\textsc{InFusE}}
\newcommand{\sentli}{\textsc{SeNtLI}}

\newcommand{\fulldoc}{\textsc{FullDoc}}
\newcommand{\summac}{\textsc{Summac}}

\usepackage{multirow}
\usepackage{graphicx}

\definecolor{highlightcolor}{rgb}{0.9, 0.9, 0.9}
\title{Leveraging Entailment Judgements in Cross-Lingual Summarisation}

\author{Huajian Zhang\thanks{~~Part of the work done for his MSc thesis at the University of Edinburgh.} \quad Laura Perez-Beltrachini \\ ILCC, School of Informatics \\
   University of Edinburgh\\
   \texttt{huajian.zhang.98@gmail.com}, \texttt{lperez@ed.ac.uk}}
\begin{document}
\maketitle
\begin{abstract}
Synthetically created Cross-Lingual Summarisation (CLS) datasets are prone to include document-summary pairs where the reference summary is unfaithful to the corresponding document as it contains content not supported by the document (i.e., hallucinated content). This low data quality misleads model learning and obscures evaluation results. Automatic ways to assess hallucinations and improve training have been proposed for monolingual summarisation, predominantly in English. For CLS, we propose to use off-the-shelf cross-lingual Natural Language Inference (X-NLI) to evaluate faithfulness of reference and model generated summaries. Then, we study training approaches that are aware of faithfulness issues in the training data and propose an approach that uses unlikelihood loss to teach a model about unfaithful summary sequences. Our results show that it is possible to train CLS models that yield more faithful summaries while maintaining comparable or better informativess.\footnote{Code and data available at \url{https://github.com/HJZnlp/Faithful_XWikis}.}
\end{abstract}

\section{Introduction}

A widely used method to create abstractive summarisation datasets is to crawl websites from which documents paired with 
reference summaries can be extracted. Examples of this are
synthetic datasets created in the news \cite{grusky-etal-2018-newsroom,narayan-etal-2018-dont,scialom-etal-2020-mlsum,hasan-etal-2021-xl} and instructional domains \cite{ladhak-etal-2020-wikilingua} and for descriptive summarisation 
\cite{liu2018generating,perez-beltrachini-etal-2019-generating,perez-beltrachini-lapata-2021-models}.
The potential content misalignment in document-summary pairs created in this way raises concerns about the quality of training and evaluation data \cite{gehrmann2022-obstacles-nlgeval}.

Previous work carried out manual validation of document-summary pairs in automatically created datasets to assess content overlap thereof
\cite{maynez-etal-2020-faithfulness,hasan-etal-2021-xl,perez-beltrachini-lapata-2021-models,gao-etal-2023-evaluating,chen-etal-2023-revisiting}.
This inspection aims at pinpointing whether summaries convey content that cannot 
be inferred from the document, i.e. \textit{hallucinations}. \cite{maynez-etal-2020-faithfulness} 
found that 70\% of the pairs in the XSum dataset \cite{narayan-etal-2018-dont} contain 
summaries with hallucinations.
For multi- and cross-lingual summarisation,
\cite{hasan-etal-2021-xl,perez-beltrachini-lapata-2021-models} found \textasciitilde30\% of the summaries to be unfaithful.

Taking a causal look into the hallucinations problem in automatically generated summaries, previous work \cite{gem-sets-2021,nan-etal-2021-entity,liu-etal-2021-noisy,cao-wang-2021-cliff,goyal-durrett-2021-annotating,choubey2022cape,aharoni2022mface,qiu2023detecting} inspects whether reference summaries in a dataset contain hallucinations in an automatic way. They further exploit this information about the data quality in terms of faithfulness for training with a better signal. Their focus is on monolingual summarisation (the input document and output summary are in the same language) and the most studied language is English. In this work, we focus on cross-lingual summarisation where the document 
is written in one language (e.g., Czech) and the corresponding summary is written in a different language (e.g., English).

We propose to leverage cross-lingual natural language inference  \cite{conneau2018xnli} to supplement human validation and to automatically annotate synthetic cross-lingual datasets with hallucination judgements. 
Our study focuses on the XWikis corpus \cite{perez-beltrachini-lapata-2021-models} consisting of descriptive summaries.
It is extracted from Wikipedia by aligning articles in different languages (e.g., Czech and English) and re-combining the lead paragraph of one article with the body of the other to form cross-lingual document-summary pairs. It includes English, French, German, and Czech and we extended it with Chinese.

We study simple training schemes that are aware of hallucinations occurring in reference summaries. 
These train with a smaller but cleaner dataset where highly unfaithful document-summary pairs are removed, include unfaithful document-summary pairs for training but ignore the unfaithful content or explicitly teach the model about unfaithful summary sub-sequences. 
Our experiments show that by simply fine-tuning with a smaller but more faithful training set it is possible to improve faithfulness on generated summaries while maintaining their informativeness in the context of CLS.

\section{X-NLI Based Faithfulness Estimation}
\label{sec:xnli-eval}

\paragraph{The XWikis Corpus}

We focus on the XWikis corpus which covers four European languages, English (en), French (fr), German (de), and Czech (cs). In addition, we follow the approach proposed in \cite{perez-beltrachini-lapata-2021-models} to extend XWikis with Chinese (zh). In Appendix~\ref{app:zh-xwikis} we show statistics about the XWikis summarisation task \cite{perez-beltrachini-lapata-2021-models} and the added zh language.
Thus, XWikis covers high resource pairs, e.g., fr-en and de-en, as well as lower resource and English-distant languages, i.e., cs-en and zh-en.  Following previous work we focus our experiments on the All-to-English language pairs.

\paragraph{Efficacy of X-NLI}

We examine the performance of existing multi-lingual NLI models in our cross-lingual setting, i.e., where the premise and hypothesis are in different languages.
We choose X-NLI \cite{conneau2018xnli} as multi-lingual NLI dataset and an off-the-shelf model, namely mT5-large, fine-tuned on the X-NLI training set, i.e., pairs of English premise and English hypothesis.\footnote{https://huggingface.co/alan-turing-institute/mt5-large-finetuned-mnli-xtreme-xnli}
The NLI model will transfer zero-shot to other languages, e.g., French, where the premise and hypothesis are in the same language \cite{conneau2018xnli}. 
The X-NLI test set includes 15 languages with a total of 5,000 premise-hypothesis pairs. 
 However, we need to verify the zero-shot performance in our cross-lingual scenario where the premise and hypothesis are in different languages. 
To this end, we derive cross-lingual premise-hypothesis pairs from the X-NLI test set. We combine the premise in one language (e.g., French) with the hypothesis in another language (e.g., English).

Table~\ref{tab:xnli} shows results for the multi- and cross-lingual scenarios. 
The performance in the cross-lingual scenario is still competitive when compared to that of monolingual English and multi-lingual ones. Thus, following previous work on monolingual English summarisation evaluation, we make use of X-NLI to assess cross-lingual summary faithfulness. Appendix~\ref{app:xnli} provides details about the X-NLI dataset and examples of cross-lingual premise-hypothesis pairs.

\begin{table}[t]
\centering
\small
\setlength{\tabcolsep}{4pt}
\def\arraystretch{1.2}
\begin{tabular}{cccccc}

\thickhline
    & en & fr &  de & cs & zh \\
en & 87.27 & 85.65 & 85.65 & -- & 84.91 \\
fr & 84.97 & 82.83 & 82.44 & -- & 81.84 \\
de & 85.81 & 83.41 & 83.05 & -- & 81.58 \\
cs &  -- & -- & -- & -- & --   \\
zh & 84.49 & 81.86 & 82.02 & -- & 81.20 \\
\thickhline
\end{tabular}
\vspace{-0.2cm}
\caption{\label{tab:xnli}
Accuracy of the X-NLI model on all language pairs of premise-hypothesis from the X-NLI test set for those languages present in XWikis.
Czech (cs) is not included in X-NLI test set.}

\end{table}
\begin{table}[h]
\centering
\small
\setlength{\tabcolsep}{4pt}
\def\arraystretch{1.2}
\begin{tabular}{lccccc}
\thickhline
Models  & fr-en & de-en & cz-en & zh-en &  AVG  \\
\hline
\fulldoc   & 61.93 & 72.87 & 63.43 & 61.45 & 64.92  \\
\textsc{Summac}$_\textsc{zs}$ & 73.42 & 80.27 & 68.34 & 65.43 & 71.87  \\
\sentli  & \textbf{80.32} & 76.38 &70.86 & 58.59 & 71.54  \\
\ours & 78.45 & \textbf{83.78} & \textbf{79.20} & \textbf{69.58} & \textbf{77.75}  \\ 
\thickhline
\end{tabular}
\vspace{-0.2cm}
\caption{\label{tab:results_eval}
Performance measured as ROC-AUC of cross-lingual NLI-based approaches on predicting faithfulness of reference summary sentences. 
}
\end{table}

\paragraph{Benchmarking NLI-based Approaches}

We benchmark existing NLI-based approaches proposed for monolingual summarisation to assess reference summary faithfulness in our cross-lingual summarisation task.
These approaches consider summary sentences as hypotheses and differ in what they consider as premise. \fulldoc~ \cite{maynez-etal-2020-faithfulness,honovich-etal-2022-true,dziri-etal-2022-evaluating} takes the entire input document as premise while \summac~ \cite{laban-etal-2022-summac} considers a single document sentence, the one that yields the highest entailment score. In a middle ground, \sentli~ \cite{schuster-etal-2022-stretching} and \ours~ \cite{infuse} consider a subset of document sentences as premise. Both rank document sentences according to the entailment score they yield when used as premise for summary sentences. \sentli~ uses a fixed number of top-$k$ ranked document sentences (together with top-$k$ highest ranked by contradiction score) as premise. In contrast, \ours~ uses a variable input number of highest entailment scoring document sentences. Further details about these approaches and our implementation can be found in Appendix~\ref{app:xnli}.

For the comparison, we use the human annotated reference summaries from the XWikis development set in \cite{perez-beltrachini-lapata-2021-models}. We use the sentence level annotations where three bilingual speakers per language pair judge whether the content conveyed by summary sentences is supported by the corresponding document. 
The benchmark contains 238 annotated sentences in total. We aggregate annotators' judgements into a single $yes$ (supported) / $no$ (unsupported) class. In Appendix~\ref{app:xwikisbenchmark}, we provide details about the annotation, aggregation and statistics.

As Table~\ref{tab:results_eval} shows, \ours~ outperforms all other approaches. We attribute this to the fact that document-summary pairs in XWikis exhibit diversity (i.e., some summary sentences are simpler or extractive requiring only a few document sentences as supportive premise, while others are complex or abstractive necessitating a larger number of document sentences). Furthermore, the main type of hallucinations in reference summaries are extrinsic ones (i.e., there is no supportive evidence in the input document) and \ours~ shows the best performance on these \cite{infuse}.

\section{Faithful Training}
\label{sec:faith_train}

\subsection{Cross-lingual Abstractive Summarisation}
\label{sec:faith_task}

We formalise cross-lingual summarisation as follows. Given an input document Doc$_{X}$ written in a language~$X$ represented as a sequence of tokens $x=(x_1 \cdots x_{|x|})$, the task is to generate a summary $\widehat{\text{Sum}}_Y$ in a target language~$Y$. $\widehat{\text{Sum}}_Y$ consists of a sequence of tokens $(y_1 \cdots y_{|y|})$ and is generated token-by-token conditioning on $x$ by a summarisation model $p_{\theta}$ as
$\prod_{t=1}^{|y|} p_{\theta}(y_t|y_{1..t-1}, x)$.

\paragraph{Automatic Faithfulness Annotation}
We split reference summaries into sentences, i.e., $\text{Sum}_Y = (y^1 \dots y^M$) where $M$ is the number of sentences. We use \ours~ to automatically annotate each reference summary sentence $y^j$ with a faithfulness judgement $F \in \{yes, no\}$. For training,  the sentence level annotations are propagated to tokens in the sentence  $y^j=(y^j_1 \cdots y^j_{|y^j|})$. Details about the annotation are provided in Appendix~\ref{app:annotation-training}.

\subsection{Faithfulness Aware Approaches}
\label{sec:faith_approaches}

We study approaches that differ in how they weight unfaithful document-summary pairs during fine-tuning. That is, approaches that range from using a hard weight where the pair is removed to applying a negative weighting scheme where the objective seeks to decrease the likelihood of unfaithful summary sequences. We propose the latter one following ideas in dialogue \cite{li-etal-2020-dont} and computer vision \cite{motion-fore-zhu22a}. All the approaches assume the initialization of the model parameters $\theta$ from a pre-trained multi-lingual language model.

\paragraph{Clean} This variant fine-tunes on a subset from the original training set. We aggregate sentence labels to the summary level and any document-summary pair labeled with $no$ will be removed. 

\paragraph{Mask}
This method sets the loss to zero for tokens in $y^j=(y^j_1 \cdots y^j_{|y^j|})$ labelled with $no$ \cite{goyal-durrett-2021-annotating}. The training loss is formulated as: 
\vspace{-0.4cm}
\begin{equation}
\mathcal{L}_{\textsc{Mask}}(\theta) = -\sum_{t=1}^{|y|} F_t \log p_\theta(y_t|x, y_{<t})
\label{eq:mask}
\end{equation}
where $F_t=0$ if $y_t$ is labelled with $no$, $1$ otherwise.

\paragraph{Unlike$_{PR}$}
We propose to use unlikelihood loss \cite{ulLoss} to fine-tune a model together with MLE. The loss term is defined as:
\vspace{-0.3cm}
\begin{align}
\mathcal{L}_{\textsc{UL}}(\theta) = 
&- \sum_{t=1}^{|y|} [
\log p_\theta(y_t|x, y_{<t}) \\
&- \alpha \sum_{c \in \mathcal{C}} \log (1-p_\theta(c|x, y_{<t}))] \label{eq:unlike}
\end{align}
where $\alpha$ is a hyperparameter for the unlikelihood weight and $\mathcal{C}$ is the set of tokens in $y^j=(y^j_1 \cdots y^j_{|y^j|})$ that are labelled with $no$. 
We further encourage more faithful training via implicit prompt, which is similar to segment embedding  \cite{devlin-etal-2019-bert}.
The intuition is, 
instead of decreasing the probability of $p(y_t|y_{<t})$ when $y_t$ is labeled as unfaithful, we learn $p(y_t|y_{<t}, F_t)$ instead to avoid impact on language modeling.  We convert $\{yes, no\}$ labels to parameterized vectors $\mathbf{p}_1$ ($yes$) and $\mathbf{p}_0$ ($no$). We integrate either $\mathbf{p}_1$ or $\mathbf{p}_0$ for model learning, while using $\mathbf{p}_1$ at inference to encourage factual sequences. 
Additionally, we augment the reference summary $\text{Sum}_Y$ with explicit prompts \texttt{<h>} and \texttt{</h>} to denote the start/end of a token sub-sequence labelled with $no$. 
At inference time, we mask out \texttt{<h>} and \texttt{</h>}.

\section{Experimental Setup}
\label{sec:faithful_eval}

We treat m\textsc{Bart50} \cite{tang2020multilingual} as our backbone model and benchmark those faithfulness aware approaches introduced in Section~\ref{sec:faith_approaches}. \textbf{Vanilla} is our baseline model fine-tuned on the original training set assuming all instances are faithful. \textbf{Random} is a variant of Clean where we remove the same number of document-summary pairs but these are randomly selected.
See details about model training in Appendix~\ref{app:annotation-training}.

\section{Evaluation}
\label{sec:eval}

Our evaluation focuses on complementary qualitative aspects, \textit{faithfulness} and \textit{informativeness}.

\paragraph{Automatic Evaluation}
To evaluate faithfulness of generated summaries $\widehat{\text{Sum}}_{en}$ given input documents Doc$_X$, we rely on the X-NLI based approach \ours~ described in Section~\ref{sec:xnli-eval}. 
For additional judgements on faithfulness, we use a monolingual metric to assess $\widehat{\text{Sum}}_{en}$ w.r.t. their corresponding input document Doc$_{en}$. Although Doc$_X$ and Doc$_{en}$ are only comparable, the intuition is that salient content will be present in Doc$_{en}$ and that this verification assesses correctness of generated content.  Note that the XWikis corpus provides cross-lingual pairs (Doc$_X$, Sum$_{en}$) but also their related document Doc$_{en}$ and reference summary Sum$_X$. Concretely, we apply UniEval \cite{unieval} which
has been shown to align better with human annotators when compared to ChatGPT-3.5 on various summarisation tasks \citet{liu2023geval}. 
UniEval frames faithfulness evaluation as a boolean question-answering task. The document and generated summary with a specific question targeting an evaluation dimension (e.g., \textit{Is this a consistent summary of the document?}) are concatenated as input. The probability of getting \textit{yes} as an answer is taken as the metric score. 
Here we only use the consistency dimension.  

Ideally, increasing the faithfulness of summaries does not come at the cost of sacrificing their informativeness. To validate this, we assess the generated against the reference summaries.
We exploit the additional multi-lingual reference summaries and compare $\widehat{\text{Sum}}_{en}$ against Sum$_{en}$ with ROUGE-L \cite{lin-2004-rouge} and against Sum$_X$ with LaBSE \cite{labse}. LaBSE relies on multi-lingual embeddings and provides cross-lingual similarity scores. 
This provides a more robust multi-reference evaluation for informativeness.
Details about metric implementations can be found in Appendix~\ref{app:evaluation}.

\paragraph{Human Evaluation}
We complemented the automatic evaluation with a human study on faithfulness and informativeness. Given that the evaluation of long document CLS  with qualified bilingual speakers is costly, we carried out the evaluation on two representative language pairs, namely fr-en which is high resource in Wikipedia and close to English and zh-en which is low resource and distant. We randomly select 20 samples for each language pair and employed two annotators per pair, with the authors serving as the third annotator. We ask the annotators to score candidate summaries on a  Likert scale from 1 (low) to 4 (high) in terms of faithfulness (\textit{Does the candidate summary contain any content that is not supported by or contradicts the given document?}) and informativeness (\textit{Does the candidate summary cover the main points in the reference summary?}).
Details about the instructions are provided in Appendix \ref{app:human_eval}.

\begin{table*}[t]
\centering
\setlength{\tabcolsep}{4pt}
\def\arraystretch{1.2}
{
\footnotesize
  \begin{tabular}{@{}c@{\hspace{0.4cm}}l@{\hspace{0.2cm}}|c@{\hspace{6pt}}c@{\hspace{6pt}}c@{\hspace{6pt}}c@{\hspace{6pt}}|c@{\hspace{6pt}}c@{\hspace{6pt}}c@{\hspace{6pt}}c@{\hspace{6pt}}@{}}
  \hline
& Model & RL & LaBSE & \ours &  UniEval & RL & LaBSE & \ours &  UniEval \\
 & & \multicolumn{4}{c|}{XWikis Test}& \multicolumn{4}{c}{XWikis Test$_{Faith}$} \\
\hline

\parbox[t]{4mm}{\multirow{5}{*}{\rotatebox[origin=m]{90}{fr-en}}}
& Vanilla   & 31.33 & 64.95& 38.14 & 55.17  & 32.71 & 67.11& 48.95 & 56.99\\
& Random    & 31.09 & 63.00&40.96 & 55.08 & 32.02 &65.07& 47.22 & 56.97\\
\cdashline{2-10}
& Clean      & 31.27 &65.20 &46.70 &56.86 & 32.93 &67.54 & 58.63 & 59.02 \\
& Mask        & 31.23 &65.11 & \textbf{47.54} & 56.69 &32.92 &67.55 &\textbf{59.34} &59.34  \\
& Unlike$_{PR}$ & \textbf{31.55} &\textbf{65.35} & 47.49   &\textbf{57.21} & \textbf{33.49} &\textbf{67.61} & 59.19   &\textbf{59.40} \\
\hline

\parbox[t]{4mm}{\multirow{5}{*}{\rotatebox[origin=m]{90}{de-en}}}
& Vanilla    & 32.15 &66.21 & 41.73 &57.60 & 34.28 &67.58 & 50.73 &59.29  \\
& Random     &32.01  & 64.23&41.28  &56.47 & 33.92 &64.27 &50.21 &56.27 \\
\cdashline{2-10}
& Clean      & 32.04 &66.21 & 43.11&58.26 & 34.09 &67.69 & 51.94&58.54 \\
& Mask       & 32.23 & 66.53 & \textbf{45.68}  &58.35  & 34.50 &68.23 & \textbf{55.01}  &\textbf{60.65}   \\
& Unlike$_{PR}$     & \textbf{32.31}  &\textbf{66.65}&  44.45 &\textbf{58.61} & \textbf{34.63}  &\textbf{68.26}&  54.19 &60.62  \\
\hline

\parbox[t]{4mm}{\multirow{5}{*}{\rotatebox[origin=m]{90}{zh-en}}}
& Vanilla     & \textbf{31.57} &64.37& 43.91&53.21 & 33.49 &67.98& 55.23&57.10 \\
& Random &30.43  &61.97 &43.11 &52.92 & 33.02&64.24&53.92& 55.36\\
\cdashline{2-10}
& Clean     &31.46&64.30&44.07&53.36 & 33.59&68.03&54.56&57.18 \\
& Mask      &31.36  &64.08& 47.85  & \textbf{54.39} & 33.45  &68.04& 57.43  & \textbf{57.38}   \\
& Unlike$_{PR}$    & 31.48 &\textbf{64.52}& \textbf{49.22} & 53.91 & \textbf{33.87} &\textbf{68.86}& \textbf{58.77} & 57.35 \\
\hline

\parbox[t]{2.8mm}{\multirow{5}{*}{\rotatebox[origin=m]{90}{cs-en}}}
& Vanilla    & \textbf{32.94} &67.33& 37.53&57.51 & 34.47 &69.01& 48.51&60.04 \\
& Random & 31.81 &64.82 & 37.36  & 57.19 & 33.25 &66.41&48.06&58.12 \\
\cdashline{2-10}
& Clean     &32.82  &67.27&39.26 &57.69 & 34.45  &69.07&49.05 &59.81  \\
& Mask      & 32.70 &67.24&39.04  & 57.45 & 34.50 &69.15&49.93  & 59.78   \\
& Unlike$_{PR}$ & 32.65 &\textbf{67.39}& \textbf{41.30}   & \textbf{58.55} &  \textbf{34.89} &\textbf{69.54}& \textbf{50.74}   & \textbf{61.89} \\
\hline

  \end{tabular}  
 }
 \vspace{-0.2cm}
\caption{Results on the XWikis test splits (the left block corresponds to the original test sets and the right block to the filtered higher faithfulness test set Test$_{Faith}$).
Average ROUGE-L (RL), LaBSE, \ours, and UniEval scores. For all metrics higher is better and the best scores are shown in bold.
}
\label{tab:comparative:results:faithful}
\end{table*}

\section{Results and Discussion}
\label{sec:results}

Table~\ref{tab:comparative:results:faithful} (left block) shows results for summarisation models fine-tuned with different signals from faithfulness judgements and evaluated on the XWikis test sets. For all language pairs, faithfulness aware approaches outperform the Vanilla variant trained on the original dataset in terms of faithfulness. 
The Clean variant outperforms the Random one across the board, showing the impact of faithfulness signal in removing training instances. As for informativeness, it is clear for high resource language pairs (fr-en and de-en) where RL and LaBSE increase. For lower resource and distant pairs (cs-en and zh-en) RL decreases slightly while LaBSE shows improvements. We speculate that test reference summaries for these language pairs contain more hallucinations. To verify this, we evaluate models in a smaller but more faithful subset from the original test set, namely Test$_{Faith}$, where we include only the highest scoring pairs in terms of \ours~ and LaBSE (i.e., reference summaries Sum$_{en}$ that best align with the input document Doc$_{X}$ and multi-lingual reference Sum$_{X}$). We provide details about the construction of Test$_{Faith}$ in Appendix~\ref{app:test:faith}. These results (Table~\ref{tab:comparative:results:faithful} right block) show a similar overall trend, faithfulness aware approaches outperform the Vanilla and Random baselines. However, all, including baselines, perform better in terms of faithfulness. Two important observations should be drawn here. One is the importance of the quality of evaluation data in the obtained results \cite{gehrmann2022-obstacles-nlgeval}. The other is, that even a model trained on noisy data as Vanilla would perform better when the input document is more informative and contains the expected details to form the summary. 
Amongst the faithfulness aware approaches, both in the original test and Test$_{Faith}$, Clean and Mask perform the closest, while Unlike$_{PR}$ performs slightly better than these. The former ones train with less signal while Unlike$_{PR}$ explicitly decreases probabilities of unfaithful summary sentences.

Table~\ref{tab:human} shows the results of our human study. Unlike$_{PR}$ (and to a lesser extent Mask) outperform the Vanilla baseline in terms of faithfulness while being comparable (zh-en) or better (fr-en) in terms of informativeness.\footnote{The Cohen Kappa inter-annotator agreement on fr-en is 0.27 for faithfulness and 0.38 for informativeness (0.15 and 0.34 on zh-en). We speculate that the low faithfulness agreement relates to annotators using different scoring ranges.}

\renewcommand{\arraystretch}{1.4}

\begin{table}[t]
\centering
\setlength{\tabcolsep}{4pt}
\def\arraystretch{1.2}
{
\footnotesize
  \begin{tabular}{@{}c@{\hspace{0.4cm}}l@{\hspace{0.2cm}}|c@{\hspace{6pt}}c@{\hspace{6pt}}@{}}
  \hline
& Model & Informativeness &   Faithfulness\\
\hline

\parbox[t]{4mm}{\multirow{4}{*}{\rotatebox[origin=m]{90}{fr-en}}}
& Vanilla   & 1.93	& 2.10
  \\
& Clean      & 1.86&
2.15  \\
& Mask        & 2.11&
2.78	 \\
& Unlike$_{PR}$ &2.26&
2.96     \\
\hline

\parbox[t]{4mm}{\multirow{4}{*}{\rotatebox[origin=m]{90}{zh-en}}}
& Vanilla      &2.30& 2.06	
\\
& Clean      &2.15 & 2.23 \\
& Mask       & 2.33 &2.61\\
& Unlike$_{PR}$     &2.36& 3.11  \\
\hline

  \end{tabular}  
 }
 \vspace{-0.2cm}
\caption{Averaged raw scores out of 4 assigned by human judges to the informativeness and faithfulness criteria on 20 samples from the XWikis test sets (fr-en and zh-en). 
}
\label{tab:human}
\end{table}

\section{Conclusion}

We propose an NLI-based approach to faithfulness evaluation of cross-lingual document-summary pairs.
We show that the cross-lingual NLI-based evaluation can approximate human judgements of summary faithfulness. We use it to automatically annotate the synthetically created XWikis corpus and show that by considering faithfulness during training models can generate more faithful summaries while maintaining informativeness.
Our cross-lingual faithfulness evaluation and faithfulness aware training approaches are relevant in the context of CLS with Large Language Models (LLMs). The evaluation is directly applicable to assess LLM generated summaries and it would be interesting to explore its use to automatically select good quality examples \cite{10.1162/tacl_a_00632} for in-context learning or fine-tuning \cite{NEURIPS2020_1457c0d6}. It would also make sense to explore the unlikelihood loss for LLM alignment \cite{li-etal-2020-dont}.

\section*{Acknowledgements}

We thank our reviewers for their constructive feedback. We are grateful to Yumo Xu and Mirella Lapata for useful feedback and discussion. We
are extremely grateful to our bilingual annotators.
Laura Perez-Beltrachini gratefully acknowledges the support of the UK Engineering and Physical Sciences Research Council (grant EP/W002876/1) and the UKRI via the Innovate UK award to Algomo Limited.

\section{Limitations}

Our experiments and conclusions are limited to  m\textsc{Bart50} \cite{tang2020multilingual}. 
It remains to be seen the effect of noisy document-summary pairs for training larger models which deliver superior performance \cite{scaling}.

In existing research that employs automatic faithfulness evaluation models to annotate document-summary pairs in a dataset, a challenge is the absence of a clear-cut threshold for deeming the pair faithful/unfaithful. Some studies \cite{aharoni2022mface} adopt a fixed threshold, such as 0.5 entailment score, where samples scoring above this value are considered faithful and those below are not. However, this one-size-fits-all threshold is not universally applicable across different datasets. Other studies \cite{qiu2023detecting} forego setting a static threshold and instead use the evaluation score directly as a weight.
With this approach it can be difficult to learn a downstream summariser if a faithfulness evaluation model scores' range over 0 to 0.7 (all training instances are penalised). Additionally, directly using the evaluation score as a weight can amplify the impact of inaccuracies in the faithfulness evaluation model. Even small differences in the evaluator model’s score, possibly caused by noise, can disproportionately affect the performance of downstream tasks.

In this work, we regard a specific percentage of the training set as faithful; to this end, we grid search a percentage that proves an adequate data quality versus quantity trade-off showing improvements on the evaluation criteria. This method is similar to a fixed threshold and is also dataset dependent.

\bibliography{xfaith}

\appendix

\section{X-NLI Based Faithfulness Estimation}
\label{app:xnli}

\begin{table*}[t]
\centering
{\small
\begin{tabular}{lcc}
Language & Premise / Hypothesis &  Label\\
\hline 
English & There's so much you could talk about on that I'll just skip that. & Entailment  \\
& I won't talk about that, even though there's a lot to cover. &  \\
\hline  
German & Es gibt so viel was ich darüber erzählen könnte, ich überspringe das einfach.  & Entailment  \\
& Ich werde nicht darüber reden, obwohl es viel zu decken gibt. &  \\
\hline  
French & Il y a tellement de choses dont vous pourriez parler que je vais juste m'en passer. & Entailment  \\
& Je n'en parlerai pas, même s'il y a beaucoup à couvrir. &  \\
\hline  

Chinese & \begin{CJK}{UTF8}{gkai}你可以讲的太多了，我就不提了。\end{CJK}& Entailment   \\
&  \begin{CJK}{UTF8}{gkai}即使需要说的很多，但我也不会谈论这个。\end{CJK}& \\
\hline

\hline 
\end{tabular}
}
\caption{Example from the X-NLI test set. The semantics of premises/hypotheses in four languages are the same. We obtain cross-lingual NLI pairs by combining the premise in one language (e.g., \textit{Il y a tellement de choses dont vous pourriez parler que je vais juste m'en passer.}) with the hypothesis in the other language (e.g., \textit{I won't talk about that, even though there's a lot to cover.}). }
\label{tab:ex-cross-xnli-prem-hyp}
\end{table*} 

\paragraph{X-NLI model}
The X-NLI \cite{conneau2018xnli} dataset was constructed in the following way. The English premises are collected from MultiNLI \cite{mnli} and corresponding English hypotheses are produced by employed workers. All premise-hypothesis pairs are further translated into 15 languages by professional translators. We derive cross-lingual premise-hypothesis pairs from the X-NLI test set by aligning the premise in one language (e.g., Chinese) with the hypothesis in another language (e.g., English). Table~\ref{tab:ex-cross-xnli-prem-hyp} shows examples of the original and derived premise-hypothesis pairs.
Note also in Table~\ref{tab:ex-cross-xnli-prem-hyp} that we do not include results for the Czech language as it is not covered in the 15 languages of the X-NLI test. Nevertheless, the performance in the cs-en cross-lingual faithfulness evaluation setting is acceptable (see Table~\ref{tab:results_eval}).

In Table~\ref{tab:xnli}, we find the performances on ${\cal D}_{X \rightarrow en}$ and ${\cal D}_{en \rightarrow X}$ are even better than the mono-lingual $D_{X \rightarrow X}$ direction in languages other than English. We guess the reason can be that the model is fine-tuned in monolingual English sentence pairs and our cross-lingual setting benefits from having premise or hypothesis sentences in English. 

\paragraph{NLI-based Faithfulness Evaluation Approaches}

\textsc{Summac}$_\textsc{ZS}$ \cite{laban-etal-2022-summac} operates under the assumption that each sentence within a summary is supported by one corresponding sentence from the document. Consequently, it selects the document sentence with the highest entailment score to serve as the context for each summary sentence.

\sentli~\cite{schuster-etal-2022-stretching} 
retrieves a fixed number of document sentences, denoted by $k$, to form its context for each summary sentence. This context is composed of the document sentences that have the highest scores for both entailment and contradiction.
Following \cite{schuster-etal-2022-stretching}, we set the value of $k=5$.

\ours~ \cite{infuse} incrementally selects entailment-ranked document sentences and evaluates summary sentences against the formed premise; if there is no further increase in the predicted neutral score, it will automatically stop yielding the entailment score of the current premise as faithfulness score. In our application of this method, we set the automatic stopping criterion as the upper bound, while requiring the premise to include at least 5 document sentences.

\section{Datasets}

\subsection{Ethics Consideration}
We honor the ACL Code of Ethics. We use XWikis under CC BY-SA 4.0 International License. XWikis includes content from Wikipedia, which is under CC BY-SA 4.0 International License and GFDL. We ensure that the data was used only for academic purposes, which aligns with the
intended use of the dataset. For data safety, it is not avoidable that some document/summary pairs can contain uncomfortable content, including hate, crimes, and wars.

For the human validation of the zh-en pair added to the XWikis corpus (see Section~\ref{app:xwikisbenchmark} and Table~\ref{tab:valid-annot}), we recruited three Chinese-English bilingual annotators with degrees in English-speaking countries colleagues in our field.  Annotators were asked to annotate 20 samples randomly extracted from the validation set. The annotation does not involve any personally sensitive information. Similarly, for the human study in Section~\ref{sec:eval} we recruited four volunteer bilingual annotators, two for fr-en and two for zh-en.

\subsection{Adding Chinese to the XWikis Corpus}
\label{app:zh-xwikis}

\begin{figure*}[t]\centering 
\begin{tabular}{c}  
\includegraphics[scale=.55]{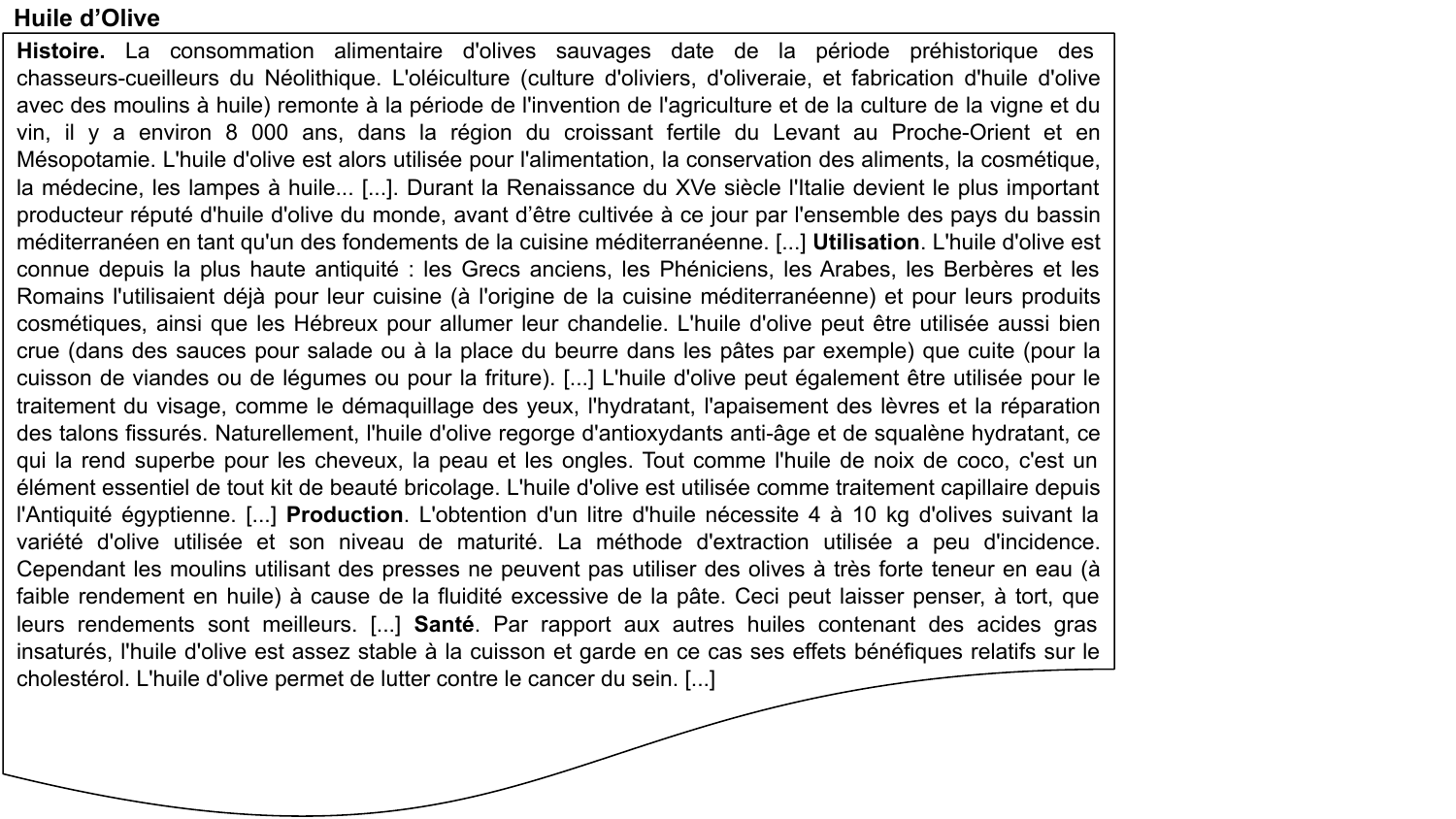} \\
\includegraphics[scale=.30]{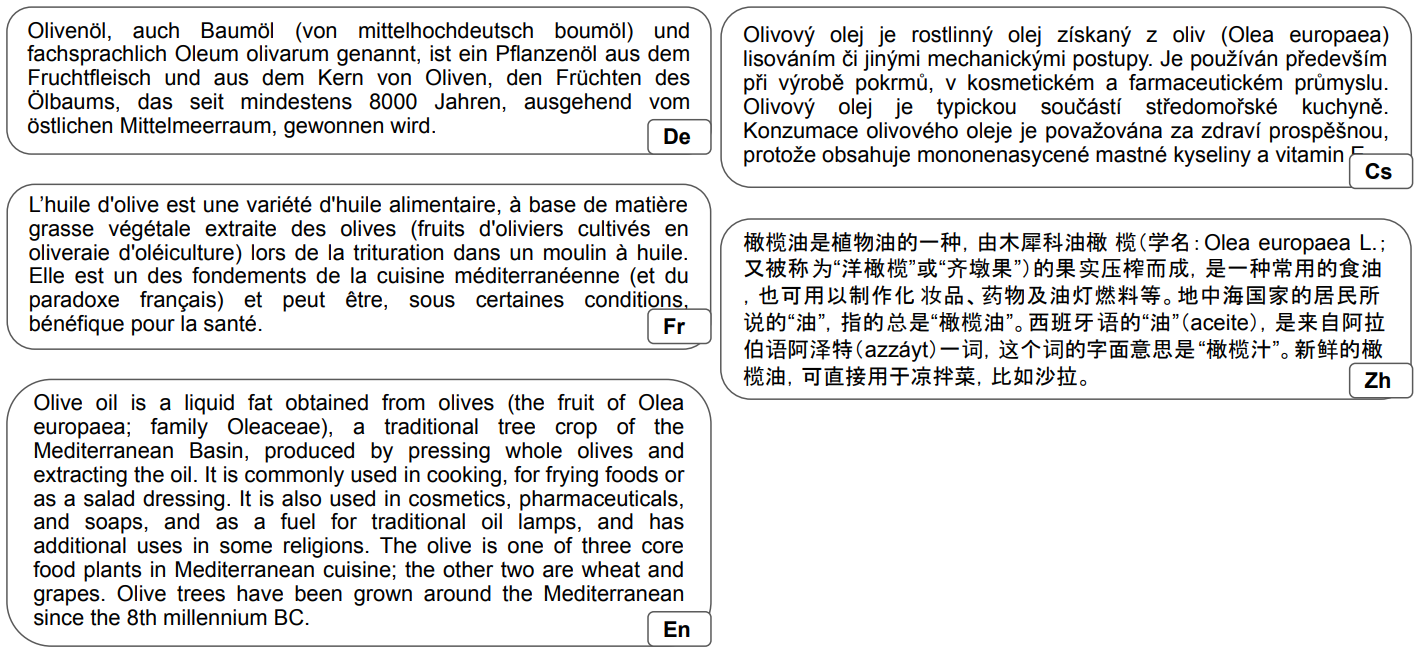}
\end{tabular} \caption{Example document in French and reference summaries in German, French, English, Czech, and Chinese.} \label{fig:clads-example} 
\end{figure*}

We follow the approach proposed in \cite{perez-beltrachini-lapata-2021-models} to extend the XWikis corpus with the Chinese language. 
Table~\ref{tab:xwikis-subsets} shows the number of instances for each language pair in the XWikis corpus including monolingual (e.g., French) subsets. Note that these are taken from the recently published version of XWikis in  the HuggingFace repository.\footnote{\url{https://huggingface.co/datasets/GEM/xwikis}}
Figure~\ref{fig:clads-example} shows an XWikis example taken from \cite{perez-beltrachini-lapata-2021-models} of document in French and corresponding reference summaries in all other languages where we include the zh summary.  

Table~\ref{tab:xwikis-stats} shows the analysis of the All-to-English XWikis CLS task carried out in \cite{perez-beltrachini-lapata-2021-models}. 
We apply the same metrics to the newly extracted Chinese subsets. 
\citet{grusky-etal-2018-newsroom} introduced metrics to evaluate how much a summary borrows textual fragments from its source document. Coverage, calculates the average number of summary tokens that are part of an extractive fragment from the document, indicating how much of the summary is directly extracted. Density, measures the average length of these extracted fragments, showing the depth of direct borrowing. Compression, calculates the ratio of the number of tokens in the summary to the number of tokens in the corresponding document. 
These metrics help assess a summary's extractiveness.
Additionally, we report \% of novel ngrams (i.e., the percentage of novel tokens that appear in the summary but are not present in the document) as an additional indication of extractiveness.
The observations drawn from this analysis for Chinese are the following. 
Chinese documents are relatively shorter than those in
the other four languages but summaries are still close in length. 
As for diversity, the aspects covered in zh are closer to those in cs, much lower than
de and fr. We speculate one reason may be due to the fewer number of document summary pairs for these languages.
Generally, zh documents indeed contain multi-topic and sections per document (Sections/Doc) match the other four languages.

\begin{table}[t]
\small
\setlength{\tabcolsep}{4pt}
\centering
\begin{tabular}{cccccc}
\hline  & en & de & fr & cs & zh \\
en & & 425,279 &468,670 & 148,519 & 135,674\\
de & 376,803 & &252,026 & 109,467& 103,044 \\
fr & 312,408 & 213,425& &91,175 & 99,301\\
cs & 64,310 & 53,275 & 51,578 && 32,588 \\
zh & 75,524& 73,969& 81,847& 43,281& \\
\hline
\end{tabular}
\caption{Total number of document-summary pairs in the XWikis corpus considering all language pairs and directions. }
\label{tab:xwikis-subsets}
\end{table}

\begin{table*}[t]
\centering
\begin{tabular}{lcccccc}
\hline & \multicolumn{4}{c}{ XWikis (comp) } &\multicolumn{2}{c}{ XWikis (Test) } \\
& de & fr & cs & zh & de/fr/cs & zh \\
\hline 
Words/Doc & 906 & 1040 & 890 &769 & 972& 890 \\
Sents/Doc & 41 & 38 & 42 &36 & 42&40 \\
Sections/Doc & 5 & 7 & 6 &6& 6&6 \\
Words/Sum & 56 & 59 & 65 & 61 & 61 & 70 \\
Sents/Sum & 3 & 2 & 3 & 3 &3&3 \\
Aspects & 253,425 & 248,561 & 65,151 &75,796 & 9,283&11,722 \\
Coverage &  65.53  &  72.23  &  55.97 &55.05 &  65.41 &62.21 \\
Density &  1.23  &  1.51  &  0.99 &0.94 &  1.23 &1.14 \\
Compression &  17.44  &  20.16  &  15.12 &15.17 &  18.35&10.89  \\
\% novel unigrams &  33.30  &  26.77  &  42.29 &39.77 &  33.25 &33.38  \\
 \quad  bigrams &  80.70  &  73.19  &  85.17  & 84.37 & 79.51&80.01  \\
 \quad  trigrams &  93.60  &  90.25  &  95.19 &95.32 &  93.17& 93.44 \\
 \quad  4-grams &  97.98  &  95.68  &  97.98 &98.13 &  97.11 &97.17 \\
LEAD &  19.09  &  23.51  &  20.21  & 12.24 & 20.88 & 13.19  \\
EXT-ORACLE &  24.59  &  28.38  &  24.25 & 16.48 &  25.95 & 17.43 \\
\hline
\end{tabular}
\caption{ XWikis statistics (number of words and sentences per document (/Doc) and summary (/Sum)) and task characterisation metrics. Statistics for de/fr/cs-en are taken from \cite{perez-beltrachini-lapata-2021-models} and zh-en are computed in this work.}
\label{tab:xwikis-stats}
\end{table*}

We also evaluate the performance of two extractive summarisation models, \textbf{LEAD} and \textbf{EXT-ORACLE}, on the validation set. We assume that these two extractive models should obtain good results if the dataset is extractive in nature. LEAD generates a summary by simply copying the first N tokens from the document, where N equals the length of the reference summaries. EXT-ORACLE is to generate a summary by selecting the portion of sentences to maximize ROUGE-2 with respect to the reference. Intuitively, when the salient information concentrates on the first several sentences of the document or so-called lead bias, LEAD performs well. And when the summarisation task is extractive, EXT-ORACLE should have a good performance \cite{perez-beltrachini-lapata-2021-models}. Rouge-L is used as the performance indicator. 

The last two rows of Table~\ref{tab:xwikis-stats} provide the results for the two extractive models. LEAD is below EXT-ORACLE by around 4 ROUGE-L points, suggesting no lead bias in the summaries. The performance of EXT-ORACLE is also not good. One reason can be that in the reference summaries, each sentence aggregates information from multiple sentences across the document. Another reason could be related to the high ratio of new paraphrases appearing in summary, which cannot be captured by an extractive model. The performance of these two models is extremely weak in Chinese. We further explore this phenomenon by translating target summaries and generated summaries (by LEAD/EXT-ORACLE) on the test set to English. They achieve 17.14 and 24.57, respectively, which can be comparable with the other three languages. One explanation is that the evaluation of ROUGE score on the other three languages is done after stemming, which is not supported for Chinese, thus leading to a drop in performance. There potentially exist other reasons such as noise introduced by the Chinese word segmentation tool in the data pre-processing stage. 

We use the following model configurations and tools for the analysis presented in Table~\ref{tab:xwikis-stats}. We apply PTBTokenizer from Stanford CoreNLP \cite{corenlp} to tokenize references. We evaluate the generated outputs with files2rouge package\footnote{https://github.com/pltrdy/files2rouge}.
For extractive methods applied to Chinese, we utilize HanLP \cite{hanlp} for segmenting sentences into words, removing stopwords, and performing tokenization.
For all other models, we use the same SentencePiece \cite{sentencepiece} model used by \cite{perez-beltrachini-lapata-2021-models}.

\subsection{Human Annotation of Hallucinations in XWikis Reference Summaries}
\label{app:xwikisbenchmark}

To assess the performance of faithfulness evaluation approaches, we use the human annotated reference summaries from the XWikis validation set in \cite{perez-beltrachini-lapata-2021-models}. We also reproduce this evaluation for the newly added Chinese subset, zh-en. It contains sentence level annotations where three bilingual speakers per language pair judge whether the content conveyed by summary sentences is supported by the corresponding input document. 
As we need a single label per sentence, we aggregate human judgements as follows. 
We map each judgement to a numerical value and distinguish when a summary sentence is partially supported, that is, $no \rightarrow 0$, $partial \rightarrow 1$, and $yes \rightarrow 2$. 
We then take the sum of the three values derived from the three annotators. If the sum is in the interval $[0-2]$ we assign a \textit{not-entail} label, if it falls in the interval $[3-6]$ we assign an \textit{entail} label. Table~\ref{tab:valid-annot} shows the percentage of \textit{not-entail}. We also show the inter-annotator agreement measured by Fleiss’s  Kappa on $yes$/$no$ judgements ($yes$ includes partial) before the aggregation (all values, except zh-en that we add here, are taken from \cite{perez-beltrachini-lapata-2021-models}). The overall moderate agreement highlights the difficulty of the task even for humans.

\begin{table}[t]
\centering
\setlength{\tabcolsep}{4pt}
\begin{tabular}{lcccc}
\hline  & de-en & fr-en &  cs-en & zh-en\\
\% not-entail  & 48.33 & 37.74 & 47.69 & 36.67  \\
Fleiss's Kappa &  0.48 &  0.55 &  0.59 &  0.47 \\
\hline
\end{tabular}

\caption{Proportion of $no$ labels, i.e., summary sentences whose content is not supported by the input document, and inter-annotator agreement.
}
\label{tab:valid-annot}
\end{table}

\subsection{Faithful Test Subsets}
\label{app:test:faith}

In order to create higher quality evaluation sets, we utilize LaBSE and \ours~ for evaluating reference English summaries from the original XWikis language pairs test sets. We aggregate their results for each sample. Those samples with top 10\% scores (700 samples) are then selected to create faithful Test$_{Faith}$ sets (each language pair has its test data). 

\section{Implementation Details}

\subsection{Automatic Faithfulness Annotation and Model Training}
\label{app:annotation-training}

We use the  m\textsc{Bart50} checkpoint provided as
mMBART-50-finetuned-many-to-many.
We follow \cite{perez-beltrachini-lapata-2021-models} setting, fine-tuning for a total of 20K updates with a batch size of 80. 
We used 2 A6000 GPUs, and the fine-tuning time cost is  45 hours.

We employ \ours~ to automatically annotate each reference summary sentence $y^j$ with a faithfulness score. We regard X\% of the sentences with the lowest score as unfaithful and leave the rest without annotations (faithful). For the Clean model variant, we average sentence level scores to sample level, and take X\% of samples with the lowest score as unfaithful. 
The determination of X for different languages was informed by an analysis of average entailment scores obtained from the cross-lingual NLI model across the five languages pairs. This inspection revealed lower entailment scores for English and Chinese, suggesting a higher prevalence of unfaithful content in summaries in these language pairs. Based on these findings, we chose a threshold value of 10 for de-en, fr-en, and cs-en; and 20 for zh-en and en-en. 
We also performed a grid search on the validation set (with X ranging from 10\% to 50\%) to confirm that this thresholds, maintain informativeness while improving faithfulness. 
Table~\ref{tab:comparative:val} shows RL and \ours~ results for the different model variants (Vanilla, Random, Clean, Mask, and Unlike$_{PR}$) on the validation set.

\renewcommand{\arraystretch}{1.4}

\begin{table}[t]
\centering
\setlength{\tabcolsep}{4pt}
\def\arraystretch{1.2}
{
\footnotesize
  \begin{tabular}{@{}c@{\hspace{0.4cm}}l@{\hspace{0.2cm}}|c@{\hspace{6pt}}c@{\hspace{6pt}}@{}}
\hline

& Model & RL &\ours  \\
\hline

\parbox[t]{4mm}{\multirow{5}{*}{\rotatebox[origin=m]{90}{fr-en}}}
& Vanilla   & 32.59 & 43.73 \\
& Random    & 32.53 & 43.05 \\
\cdashline{2-4}
& Clean     & 32.79 & 47.21 \\
& Mask      & 33.01 & \textbf{49.25} \\
& Unlike$_{PR}$ & \textbf{33.22} & 48.77 \\
\hline

\parbox[t]{4mm}{\multirow{5}{*}{\rotatebox[origin=m]{90}{de-en}}}
& Vanilla    & 33.24 & 43.54 \\
& Random     & 33.05 & 43.42 \\
\cdashline{2-4}
& Clean      & 33.14 & 46.40 \\
& Mask       & 33.32 & \textbf{48.90} \\
& Unlike$_{PR}$ & \textbf{33.64} & 47.44 \\
\hline

\parbox[t]{4mm}{\multirow{5}{*}{\rotatebox[origin=m]{90}{zh-en}}}
& Vanilla    & \textbf{32.52} & 43.53 \\
& Random     & 31.50 & 43.31 \\
\cdashline{2-4}
& Clean      & 32.09 & 44.12 \\
& Mask       & 32.44 & 50.20 \\
& Unlike$_{PR}$ & 32.36 & \textbf{50.98} \\
\hline

\parbox[t]{2.8mm}{\multirow{5}{*}{\rotatebox[origin=m]{90}{cs-en}}}
& Vanilla    & \textbf{34.29} & 38.16 \\
& Random     & 34.14 & 37.53 \\
\cdashline{2-4}
& Clean      & 34.23 & 38.79 \\
& Mask       & 34.15 & 41.09 \\
& Unlike$_{PR}$ & 34.23 & \textbf{41.44} \\
\hline

\parbox[t]{2.8mm}{\multirow{5}{*}{\rotatebox[origin=m]{90}{en-en}}}
& Vanilla    & 35.62 & 44.74 \\
& Random     & 35.43 & 44.16 \\
\cdashline{2-4}
& Clean      & 35.58 & 46.44 \\
& Mask       & 35.07 & 47.93 \\
& Unlike$_{PR}$ & \textbf{35.69} & \textbf{48.63} \\
\hline

  \end{tabular}  
 }
 \vspace{-0.2cm}
\caption{Results on the XWikis validation splits.
Average ROUGE-L (RL) and \ours~scores. 
}
\label{tab:comparative:val}
\end{table}

\begin{table}[h]
\centering
\setlength{\tabcolsep}{4pt}
\def\arraystretch{1.2}
{
\footnotesize
  \begin{tabular}{@{}c@{\hspace{0.4cm}}l@{\hspace{0.2cm}}|c@{\hspace{6pt}}c@{\hspace{6pt}}c@{\hspace{6pt}}c@{\hspace{6pt}}@{}}
  \hline
& Model & RL & LaBSE & \ours &  UniEval \\
\hline
\parbox[t]{2mm}{\multirow{5}{*}{\rotatebox[origin=m]{90}{en-en}}}
& Vanilla    & 31.12 &68.32&49.21 &78.48  \\
& Random & 30.94&65.87&48.58&75.06 \\
\cdashline{2-6}
& Clean   & 31.21 &\textbf{68.48}&51.21&\textbf{79.71}   \\

& Mask      & 30.72 &68.14& 51.38 &78.72    \\
& Unlike$_{PR}$     & \textbf{31.28} &68.43& \textbf{51.98}&78.91    \\
\hline

 \end{tabular}
 }
\caption{Results on the XWikis en-en test split.
Average ROUGE-L (RL), LaBSE, \ours, and UniEval scores. 
}
\label{tab:comparative:results:enen}
\end{table}

\begin{table*}[t]
\centering
\small
\setlength{\tabcolsep}{4pt}
\begin{tabular}{cccccccccccc}
\hline
Ratio & RL &LaBSE & \ours &  UniEval   & Coverage & Density & Compression & unigrams & bigrams & trigrams & 4grams \\ 
\hline
10\% & 31.06 & 68.00&47.69&78.09  & 0.83 & 5.43 & 12.93 & 0.17 & 0.49 & 0.66 & 0.73 \\ 
20\% & \textbf{31.21} & \textbf{68.48}&51.21 &79.71 & 0.83 & 6.51 & 12.45 & 0.17 & 0.47 & 0.63 & 0.70 \\ 
30\% & 30.88 &67.79& 50.99 & 77.91 & 0.83 & 4.83 & 13.03 & 0.17 & 0.49 & 0.67 & 0.75 \\ 
40\% & 30.68 &68.36& \textbf{55.44} & \textbf{80.95} & 0.85 & 8.32 & 11.78 & 0.15 & 0.43 & 0.58 & 0.65 \\ 
50\% & 30.69 &68.09& 55.21 & 80.58 & 0.85 & 7.40 & 12.29 & 0.15 & 0.44 & 0.59 & 0.66 \\ 
\hline
\end{tabular}
\caption{\label{tab:clean}
Results on the en-en  XWikis test split for the Clean approach. Ratio shows the percentage of removed data examples from the training set. 
Average INFUSE,
ROUGE-L (RL), LaBSE, and UniEval scores are the evaluation metrics introduced in Section~\ref{sec:eval}. The other metrics are those extractive metrics described in Section~\ref{app:zh-xwikis}.
}
\end{table*}

\subsection{Automatic Evaluation}
\label{app:evaluation}

LaBSE \cite{labse} is a dual-encoder model based on the pre-trained BERT model. It is further fine-tuned on translation ranking tasks.
It is trained on mono-lingual and bi-lingual data. The monolingual data contains CommonCrawl4
and Wikipedia. The translation corpus is constructed from Web pages using a bitext
mining system \cite{bitext}. 
We utilize its PyTorch implementation available at \url{https://github.com/yang-zhang/labse-pytorch}.
For \ours~ \cite{infuse}, we access the code from \url{https://github.com/HJZnlp/infuse}. 
For UniEval \cite{unieval}, we employ the implementation at \url{https://github.com/maszhongming/UniEval}, specifically downloading UniEval-Sum, which is designed for evaluating summarisation tasks. UniEval reports evaluation scores across four dimensions, but we only focus on consistency, which measures the factual alignment between a summary and the corresponding document.

\section{Additional Results and Analyses}
\label{app:add:analyses}

\subsection{The Effect of Faithfulness Aware Training in Monolingual English Summarisation}

Additionally, we report results for the monolingual English subset (en-en) from XWikis in Table~\ref{tab:comparative:results:enen}. Interestingly, in this setting it is also possible to train with faithfulness judgements on document-summary pairs and obtain models that generate more faithful summaries without loss in informativeness.
As this is a monolingual summarisation task, it enables us to inspect the relation between performance improvements in terms of faithfulness and informativeness, amount of training data, and extractiveness of generated summaries.
Table~\ref{tab:clean} shows statistics about performance versus extractiveness for the Clean approach for different amounts of annotated document-summary pairs considered (i.e., removed). 
For informativeness metrics, RL and LaBSE, the optimal performance is observed upon the removal of 20\% of noisy (i.e., with hallucinations) document-summary pairs, whereas, for faithfulness metrics, \ours~ and UniEval, the performance generally improves as the ratio increases up to 40\%. 
We can also observe that summaries generated by Clean variants fine-tuned with less but higher faithfulness scoring document-summary pairs tend to become more extractive (i.e., incorporating fewer novel n-grams). A more extractive summary, which largely copies content directly from the document, is indeed more faithful but potentially less informative. 
This observation suggests that although simply filtering out noisy document-summary pairs can lead to better summaries in terms of both faithfulness and informativeness, it is crucial to maintain a balance between the two \cite{ladhak-etal-2022-faithful}.

\subsection{Ablation Experiments on the Unlike$_{PR}$ Model Variant}
\label{app:ablation:unlike}

We further inspect the role of segment embeddings in Unlike$_{PR}$. We remove segment embeddings from Unlike$_{PR}$ and name the resulting model \textbf{Unlike}. Table~\ref{tab:ablation} shows the performance of Unlike$_{PR}$ versus Unlike. We observe that Unlike$_{PR}$ outperforms Unlike in terms of informativeness (RL and LaBSE). For faithfulness, \ours~ and UniEval, in low-resource (smaller training sets and data in the base pre-trained language model) language pairs (cs-en and zh-en), Unlike$_{PR}$ also performs better. However, in high-resource languages (de-en and fr-en), Unlike obtains higher \ours~ scores (higher than all approaches in Table~\ref{tab:comparative:results:faithful}). One possible reason for this is that fr-en and de-en are far larger datasets and thus they receive a stronger signal to adhere to the input document from Unlike.


\renewcommand{\arraystretch}{1.4}
\begin{table}[t]
\centering
\setlength{\tabcolsep}{4pt}
\def\arraystretch{1.2}
{
\footnotesize
  \begin{tabular}{@{}c@{\hspace{0.4cm}}l@{\hspace{0.2cm}}|c@{\hspace{6pt}}c@{\hspace{6pt}}c@{\hspace{6pt}}c@{\hspace{6pt}}@{}}
  \hline
& Model & RL & LaBSE & \ours &  UniEval \\
\hline

\parbox[t]{4mm}{\multirow{2}{*}{\rotatebox[origin=c]{90}{fr-en}}}

& Unlike &  31.37&64.22	&\textbf{48.30}& 56.77      \\
& Unlike$_{PR}$ & \textbf{31.55} &\textbf{65.35} & 46.49   &\textbf{57.21}      \\
\hline

\parbox[t]{4mm}{\multirow{2}{*}{\rotatebox[origin=c]{90}{de-en}}}

& Unlike &  32.20&	65.14&\textbf{45.53} &57.33    \\
& Unlike$_{PR}$     & \textbf{32.31}  &\textbf{66.65}&  44.45 &\textbf{58.61}  \\
\hline

\parbox[t]{4mm}{\multirow{2}{*}{\rotatebox[origin=c]{90}{zh-en}}}
& Unlike &  31.09 &63.90&49.03 & 53.10    \\
& Unlike$_{PR}$    & \textbf{31.48} &\textbf{64.52}& \textbf{49.22} & \textbf{53.91 }    \\
\hline

\parbox[t]{3mm}{\multirow{2}{*}{\rotatebox[origin=c]{90}{cs-en}}}

& Unlike & 32.03&65.78&	40.10  &57.65   \\
& Unlike$_{PR}$ & \textbf{32.65} &\textbf{67.39}& \textbf{41.30}   & \textbf{58.55}    \\
\hline

  \end{tabular}  
 }
 \vspace{-0.2cm}
\caption{Results on the XWikis test splits.
Average ROUGE-L (RL), LaBSE, \ours, and UniEval scores. 
}
\label{tab:ablation}
\end{table} 

\subsection{Faithfulness in Out-of-distribution Summarisation}
\label{app:cross}

We additionally evaluate the faithfulness aware training approaches in a cross-lingual out-of-distribution setting. This experiment aims at assessing whether models that are trained in higher quality (i.e., more faithful) data will exhibit better faithfulness when evaluated in out-of-distribution. To this end, we follow \cite{perez-beltrachini-lapata-2021-models} and evaluate zero-shot the different model variants trained in XWikis monolingual English over the Voxeurop dataset on cross-lingual news summarisation. Results are shown in Table~\ref{tab:voxeurop}. 

Zero-shot results are generally lower than those shown in Table~\ref{tab:comparative:results:faithful}, indicating that zero-shot performance remains a challenge. In terms of faithfulness (\ours~ and UniEval), the Unlike$_{PR}$ method significantly outperforms others. However, regarding informativeness (RL and LaBSE), all model variants show small differences.
Models trained on Wikipedia can capture the gist of news documents and all faithfulness aware variants show good transfer ability. However, the summary patterns differ between Wikipedia and the news domain. Consequently, these models achieve similarly low scores on metrics requiring reference summaries (RL and LaBSE).


\begin{table}[t]
\centering
\setlength{\tabcolsep}{4pt}
\def\arraystretch{1.2}
{
\footnotesize
  \begin{tabular}{@{}c@{\hspace{0.4cm}}l@{\hspace{0.2cm}}|c@{\hspace{6pt}}c@{\hspace{6pt}}c@{\hspace{6pt}}c@{\hspace{6pt}}@{}}
  \hline
& Model & RL & LaBSE & \ours &  UniEval \\
\hline

\parbox[t]{2mm}{\multirow{4}{*}{\rotatebox[origin=m]{90}{fr-en}}}

& Vanilla & 20.62 & \textbf{41.84} &  31.30& 53.90 \\
& Random    & 20.21&41.09&30.83&53.57\\
\cdashline{2-6}
& Clean & 20.51 & 41.12 &  31.54& 55.14 \\
& Mask & 20.82 &41.65  &  39.48& 57.98 \\
& Unlike$_{PR}$ & \textbf{20.95} &41.03  &  \textbf{42.69} & \textbf{62.50} \\

\hline

\parbox[t]{2mm}{\multirow{4}{*}{\rotatebox[origin=m]{90}{de-en}}}
&Vanilla & 21.14 & 42.16 &  31.86& 52.88 \\
& Random    & 20.85&41.70&31.06&52.29\\
\cdashline{2-6}
&Clean & 21.10 & 42.08 & 34.68& 55.00 \\
&Mask & 21.14 &\textbf{42.55}  &38.45  & 55.21 \\
&Unlike$_{PR}$ & \textbf{21.42} & 41.98& \textbf{ 44.70} & \textbf{60.34} \\
\hline

\parbox[t]{1mm}{\multirow{4}{*}{\rotatebox[origin=m]{90}{cs-en}}}
&Vanilla & 21.23 & 44.85 &  35.00& 52.41 \\
& Random    &20.10&44.35&34.19&52.91 \\
\cdashline{2-6}
&Clean & 20.51 &\textbf{44.88} & 39.26&54.68  \\
&Mask & 20.51 & 44.67&  39.32 & 55.31 \\
&Unlike$_{PR}$ & \textbf{21.60} & 44.52 & \textbf{45.15} & \textbf{59.12} \\
\hline
  \end{tabular}  
 }
 \vspace{-0.2cm}
\caption{Results on the Voxeurop test splits. All models are trained in monolingual english setting.  
Average ROUGE-L (RL), LaBSE, \ours, and UniEval scores. 
}
\label{tab:voxeurop}
\end{table}

\section{Human evaluation}
\label{app:human_eval}

We conduct a human evaluation to verify the effectiveness of faithfulness-aware approaches. To this end, we hire three bilingual annotators for fr-en and two for fr-en. We present them with 20 document-summary samples of the corresponding language pair. Each sample includes a document and four summaries generated by the Vanilla, Clean, Mask, and Unlike$_{PR}$ models. The annotators are tasked to score each summary first in terms of faithfulness and then in terms of informativeness. The scores are on a Likert scale from 1 (low) to 4 (high).
The full instructions are shown below. 

\paragraph{Annotation Instructions}

To run this annotation task you should be native (or near native) of French (Chinese) and have a good level of English.

A cross-lingual summary expresses relevant information from a document in a different language than the language of the document. For instance, an English summary from the French (Chinese) document for the Wikipedia title Olive Oil is shown in the example below.\footnote{The French (Chinese) and Englih document-summary pair as the one in Figure~\ref{fig:clads-example} was shown in the interface; together with example annotated candidate summaries.} Your task is to assess the quality of cross-lingual summaries in terms of content overlap with the underlying document. To judge this, you will read a document and corresponding candidate summaries. Then, you will score the candidate summaries in terms of consistency (low consistency 1 to high consistency 4) and then score them in terms of informativeness (low informativeness 1 to high 4). 

\paragraph{Criteria} 

\textit{Consistency}: Does the candidate summary contain any content that is not supported by or contradicts the given document? You should \textbf{ONLY} refer to the document when evaluating this aspect.

\textit{Informativeness}: Does the candidate summary cover the main points in the reference summary? (That is, which candidate summary covers better/more content expressed in the reference). To respond to this question you should \textbf{ONLY} look at the reference summary.

\section{Examples of Model Outputs}

Tables~\ref{tab:mbart50_outputs_fr}, \ref{tab:mbart50_outputs_cs}, \ref{tab:mbart50_outputs_zh}, and \ref{tab:mbart50_outputs_de} show examples of the baseline (Vanilla) and the studied faithfulness aware approaches. Hallucinated text spans are highlighted in gray. We manually verify their content against the corresponding documents. We can see that across all language pairs Unlike$_{PR}$ has fewer hallucinated spans.



\begin{table*}[htbp]
\centering
\small
\begin{tabular}{@{}p{0.1\linewidth}p{0.85\linewidth}@{}}
\toprule
\textbf{Model} &  
\\
\midrule
Reference & 
The canton of St. Gallen, also canton of St Gall ( ; dialectally , , , ), is a canton of Switzerland. The capital is St. Gallen.
\\
Vanilla & The canton of Saint-Gall () is a canton in Switzerland. It is located in the north-eastern part of Switzerland and is bordered by the Rhine valley to the \colorbox{highlightcolor}{south} by the cantons of Grisons, Glaris and Schwytz. The population of the canton is . \\
Clean & The canton of Saint-Gall () is an administrative division of Switzerland. It is located in the northeastern part of the state of \colorbox{highlightcolor}{Saxony}. The administrative divisions of the canton are: Its seat is in St. Gallen. \\
Mask &The canton of Saint-Gall () is a canton in Switzerland. It was created on 19 February 1803. Its territory covers the following territories: It is bordered on the north by Lake Constance and Thurgovia. On the east it is separated from Austria and Liechtenstein by the Rhine Valley, on the south by the cantons of Grisons, Glaris and Schwytz and on the west by Zurich. \\
Unlike$_{PR}$ & The canton of Saint-Gall () is an administrative division of Switzerland. It is located in the north-eastern part of the country. The administrative center of the canton is the city of St. Gallen. \\
\bottomrule
\end{tabular}
\caption{Example comparing the Reference summary with the generated ones by the baseline (Vanilla) and the faithfulness aware approaches (Clean, Mask, and Unlike$_{PR}$) from the fr-en test set.\label{tab:mbart50_outputs_fr}}
\end{table*}




\begin{table*}[htbp]
\centering
\small
\begin{tabular}{@{}p{0.1\linewidth}p{0.85\linewidth}@{}}
\toprule
\textbf{Model} &  
\\
\midrule
Reference & 
Piaget SA () is a Swiss luxury watchmaker and jeweller. Founded in 1874 by Georges Piaget in the village of La Côte-aux-Fées, Piaget is currently a subsidiary of the Swiss Richemont group.

\\
Vanilla & \colorbox{highlightcolor}{La Côte-aux-Fées} () is a Swiss luxury watch manufacturer. Founded in 1874 by Georges Edouard Piaget, the company produces watches and jewellery. The company's headquarters are in \colorbox{highlightcolor}{Neuchâtel}, Switzerland. It is one of the world's best-selling luxury brands. \\
Clean & Piaget is a Swiss luxury watch manufacturer based in La Côte-aux-Fées, Switzerland. The company was founded in 1874 by Georges Edouard Piaget, \colorbox{highlightcolor}{son of Georges Piaget}. It is best known for its watches and watchmaking machinery. Piaget's watches are made in a variety of shapes and sizes.
 \\
Mask &Piaget is a Swiss luxury watch manufacturer based in La Côte-aux-Fées. Founded in 1874 by Georges Edouard Piaget, the company produces watches and jewellery. \colorbox{highlightcolor}{The company's name is derived from the French word "Piaget", which means "horsemaker".}
 \\
Unlike$_{PR}$ & Piaget is a Swiss luxury watch manufacturer based in the village of La Côte-aux-Fées. Founded in 1874 by Georges Edouard Piaget, the company produces watches and jewellery. The company is known for its high precision watches, including the "Emperador" watches. \\
\bottomrule
\end{tabular}
\caption{Example comparing the Reference summary with the generated ones by the baseline (Vanilla) and the faithfulness aware approaches (Clean, Mask, and Unlike$_{PR}$) from the cz-en test set.\label{tab:mbart50_outputs_cs}}
\end{table*}




\begin{table*}[htbp]
\centering
\small
\begin{tabular}{@{}p{0.1\linewidth}p{0.85\linewidth}@{}}
\toprule
\textbf{Model} &  
\\
\midrule
Reference & Giorgio Vasari (, , ; 30 July 1511 – 27 June 1574) was an Italian painter, architect, writer, and historian, best known for his "Lives of the Most Excellent Painters, Sculptors, and Architects", considered the ideological foundation of art-historical writing. He was also the first to use the term "Renaissance" in print.\\
Vanilla & Giorgio Vasari (30 July 1511 – 27 June 1574) was an Italian painter, sculptor, and architect. He is best known for his portraits of \colorbox{highlightcolor}{the walls and ceilings} of the Palazzo della Cancelleria in \colorbox{highlightcolor}{Florence}, Italy, and for his murals in Rome. \\
Clean & Giorgio Vasari (30 July 1511 – 27 June 1574) was an Italian painter, painter and painter of the Renaissance. He is best known for his portraits of \colorbox{highlightcolor}{the walls and ceilings} of the Palazzo Vecchio, the Palazzo della Cancelleria, and the Palazzo dei Cento Giorni. \\
Mask & Giorgio Vasari (30 July 1511 – 27 June 1574) was an Italian painter, sculptor, and art theorist. He is best known for his paintings \colorbox{highlightcolor}{on the walls and ceilings} of the Palazzo della Cancelleria in \colorbox{highlightcolor}{Florence}, and for the \colorbox{highlightcolor}{frescoes} in the Palazzo Vecchio in \colorbox{highlightcolor}{Rome}.  \\
Unlike$_{PR}$ & 
Giorgio Vasari (30 July 1511 – 27 June 1574) was an Italian Renaissance painter and sculptor. He is best known for his frescoes in the Palazzo della Cancelleria (Palazzo dei Cento Giorni) in \colorbox{highlightcolor}{Florence} and in Rome, as well as for his work on the walls and ceilings of the Palazzo Vecchio.\\
\bottomrule
\end{tabular}
\caption{Example comparing the Reference summary with the generated ones by the baseline (Vanilla) and the faithfulness aware approaches (Clean, Mask, and Unlike$_{PR}$) from the zh-en test set.\label{tab:mbart50_outputs_zh}}

\end{table*}




\begin{table*}[htbp]
\centering
\small
\begin{tabular}{@{}p{0.1\linewidth}p{0.85\linewidth}@{}}
\toprule
\textbf{Model} &  
\\
\midrule
Reference & 
Rudolf Karl Bultmann (; 20 August 1884 – 30 July 1976) was a German Lutheran theologian and professor of the New Testament at the University of Marburg. He was one of the major figures of early-20th-century biblical studies. A prominent critic of liberal theology, Bultmann instead argued for an existentialist interpretation of the New Testament. His hermeneutical approach to the New Testament led him to be a proponent of dialectical theology.

\\
Vanilla & Rudolf Bultmann (\colorbox{highlightcolor}{November 17, 1855 – January 22, 1951}) was a German Protestant theologian and philosopher. He was born in Oldenburg, Germany. His father was a liberal theologians, and his mother was a pietist. \\
Clean & Rudolf Bultmann (\colorbox{highlightcolor}{February 1, 1855 – March 2, 1951}) was a German Protestant theologian and philosopher. He was born in Oldenburg, Germany, and died in Marburg, Germany. His father was a \colorbox{highlightcolor}{Lutheran} pastor.
 \\
Mask &Rudolf Bultmann (\colorbox{highlightcolor}{October 22, 1855 – February 26, 1951}) was a German Protestant theologian and philosopher. He was born in Oldenburg, Germany. His father was a liberal theologians, and his mother was a pietist. In 1895 he attended the \colorbox{highlightcolor}{gymnasium} at Oldenburg. During this time he was a member of the student association Camera obscura Oldenburgensis.
 \\
Unlike$_{PR}$ & Rudolf Bultmann (born Arthur Kennedy Bultman in Oldenburg, Germany; died in Marburg, Germany) was a German Protestant theologian and philosopher. He is best known for his work on the New Testament and its mythology, which he published in 1921.
 \\
\bottomrule
\end{tabular}
\caption{Example comparing the Reference summary with the generated ones by the baseline (Vanilla) and the faithfulness aware approaches (Clean, Mask, and Unlike$_{PR}$) from the de-en test set. 
\label{tab:mbart50_outputs_de}}

\end{table*}


\end{document}